\let\accentvec\vec

\documentclass{svproc}
 
     \let\vec\accentvec
     
\usepackage{graphics}
\usepackage{diagbox}
\usepackage{epsfig}
\usepackage{mathtools}
\usepackage{times}
\usepackage{amsmath}
\usepackage{amssymb}
\usepackage{xcolor}
\usepackage[english]{babel}
\usepackage[utf8x]{inputenc}
\usepackage[T1]{fontenc}
\usepackage{lmodern}
\usepackage{graphicx}
\usepackage[english]{babel}
\usepackage{algorithm}
\usepackage[noend]{algpseudocode}
\usepackage{support-caption}
\usepackage{subcaption}
\usepackage[colorinlistoftodos]{todonotes}
\usepackage[colorlinks=true, allcolors=blue]{hyperref}

\usepackage{silence}
\begin{document}
\mainmatter              
\title{Illumination-invariant Face recognition by fusing thermal and visual images via gradient transfer}
\titlerunning{Illumination-invariant Face recognition}
\author{Sumit Agarwal , Harshit S. Sikchi, Suparna Rooj, Shubhobrata Bhattacharya and Aurobinda Routray}
\authorrunning{Sumit et al.}
\institute{Indian Institute of Technology, Kharagpur\\
}
\makeatletter
\edef\orig@output{\the\output}
\output{\setbox\@cclv\vbox{\unvbox\@cclv\vspace{0pt plus 20pt}}\orig@output}
\makeatother

\maketitle              

\begin{abstract}
Face recognition in real life situations like low illumination condition is still an open challenge in biometric security. It is well established that the state-of-the-art methods in face recognition provide low accuracy in the case of poor illumination. In this work, we propose an algorithm for a more robust illumination invariant face recognition using a multi-modal approach. We propose a new dataset consisting of aligned faces of thermal and visual images of a hundred subjects. We then apply face detection on thermal images using the biggest blob extraction method and apply them for fusing images of different modalities for the purpose of face recognition. An algorithm is proposed to implement fusion of thermal and visual images. We reason for why relying on only one modality can give erroneous results. We use a lighter and faster CNN model called MobileNet for the purpose of face recognition with faster inferencing and to be able to be use it in real time biometric systems. We test our proposed method on our own created dataset to show that real-time face recognition on fused images shows far better results than using visual or thermal images separately.

\end{abstract}

\begin{figure}[ht]
\centering
\includegraphics[width=0.7\textwidth]{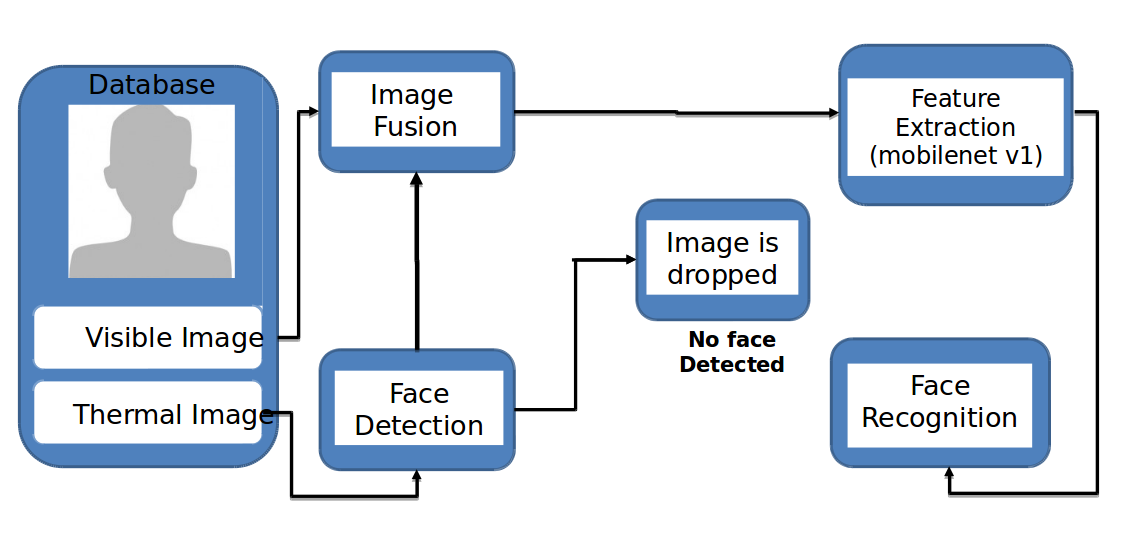}
\caption{Flowchart}
\label{fig:flowdiagram}
\end{figure}

\section{Introduction}
In this age of smart technologies, biometric plays an important role in keeping us secure. Devices that use face recognition as biometric are non-intrusive, reliable and convenient. Face recognition is considered to be best suited for identification \cite{ekenel2007face}, \cite{pentland2000face}, \cite{pentland2000face}. Since, the pioneer work of \cite{galton1889personal} a number of research work has been proposed like \cite{belhumeur1997eigenfaces}, \cite{he2005face}, \cite{gao2002face}, \cite{kirby1990application}, and \cite{bartlett2002face} but some unsolved problems still persists. The performance of such algorithms are vulnerable to poor illumination condition \cite{adini1997face}, disguises and spoofing attacks \cite{wen2015face}. Literature evince that Infrared Imaging has the potential to counter-attack the aforementioned problems \cite{cutler1996face}, \cite{bebis2006face}, \cite{socolinsky2003face}. However, thermal imagery has its own drawbacks – it is opaque to glasses, sensitive to surrounding temperature, the distribution of heat changes with individual’s facial expression. Also, face detection and holistic feature extraction is still a challenge from thermal images \cite{forczmanski2016human}, \cite{wong2012face}. 

One of the objectives of this paper is to propose an efficient face recognition algorithm against the mentioned constraints. The proposed method uses the positive characteristics of both visible and thermal spectra to recognize a given face. The limitations of both the domain are addressed by fusing the thermal and visible images optimally to get a reliable result. Our experimental results reinforce our claim of solving the problem of losing eyes information under spectacles, poor recognition under poor illumination. Also, thermal maps cannot be generated artificially by playing video or showing images to the biometric system which can be done with the intention of breaching the security. 

A database was needed to test the performance of our methodology. While a vast number of databases designed for various tasks exist for the visual spectrum, only a few relevant thermal face databases prevail. In the past, the most prominent databases used for facial image processing on the thermal infrared domain were the EQUINOX HID Face Database \cite{selinger2006appearance} and the IRIS databases. The NIST/Equinox database contains image pair’s co-registered using hardware setting. The image pairs in the UTK-IRIS Thermal/Visible Face Database are not, and therefore spatial alignment is required before fusion. However, both resources are no longer available. Kotani Thermal Facial Expression (KTFE) Database \cite{nguyen2013thermal} is another such database, but it is small and consists of limited examples. A currently available database upon request is the Natural Visible and Infrared facial Expression database (USTC-NVIE) \cite{wang2010natural}. The database is multimodal, containing both visible and thermal videos acquired simultaneously. The spatial resolution of the infrared videos is $320 \times 240$ pixels. However, both the thermal and visual images were captured at different angles which makes manual annotation difficult as the images are not in the same orientation. To fulfill the need for such a dataset, we developed a simultaneous thermal and visual face dataset. The dataset contains hundred subjects, whose thermal and visual images are taken simultaneously to avoid the face alignment issues. 

In this paper: Section 1: explains the proposed dataset and its protocol and also shows few samples; Section 2: contains the explanation of the procedure used in this paper which are subdivided into face detection part in thermal imagery, visible and thermal image fusion part with its optimization and the face recognition part; Section 3: shows the experimental results and accuracy obtained using proposed method and; Section 4: concludes the paper. The methodology is also presented in the Figure \ref{fig:flowdiagram}.

\subsection{Proposed Dataset}
\begin{figure}
   
    \centering
    \begin{subfigure}[b]{0.3\textwidth}
        \includegraphics[width=\textwidth]{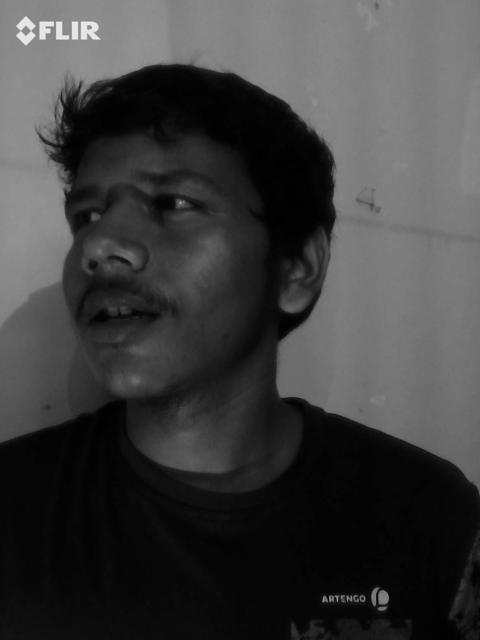}
        \label{fig:14v}
    \end{subfigure}
    \begin{subfigure}[b]{0.3\textwidth}
        \includegraphics[width=\textwidth]{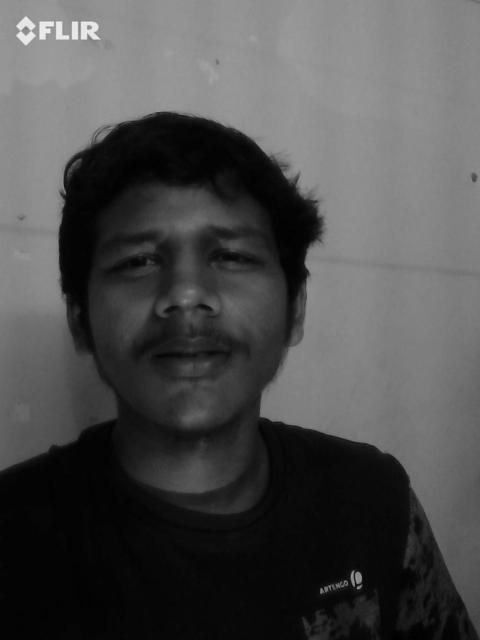}
        \label{fig:13v}
    \end{subfigure}
    \begin{subfigure}[b]{0.3\textwidth}
        \includegraphics[width=\textwidth]{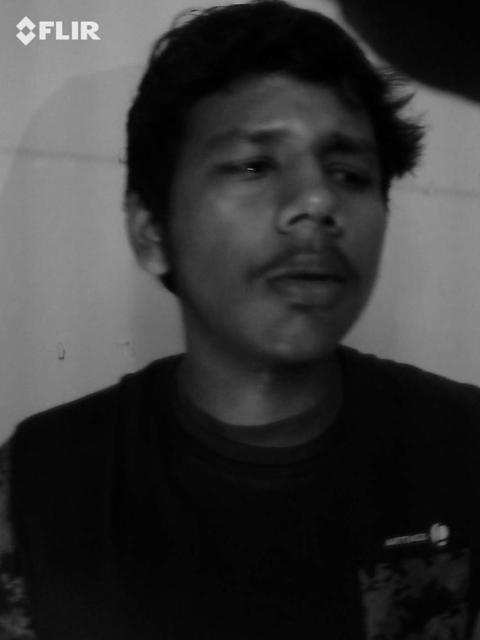}
        \label{fig:10v}
    \end{subfigure}
    \\
    \begin{subfigure}[b]{0.3\textwidth}
        \includegraphics[width=\textwidth]{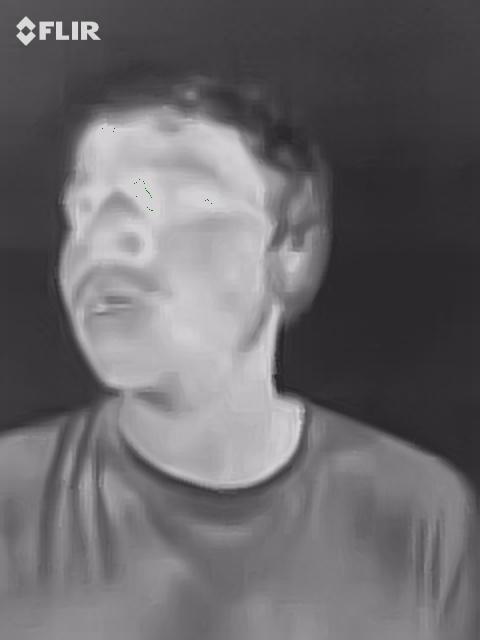}
        \label{fig:14}
    \end{subfigure}
    \begin{subfigure}[b]{0.3\textwidth}
        \includegraphics[width=\textwidth]{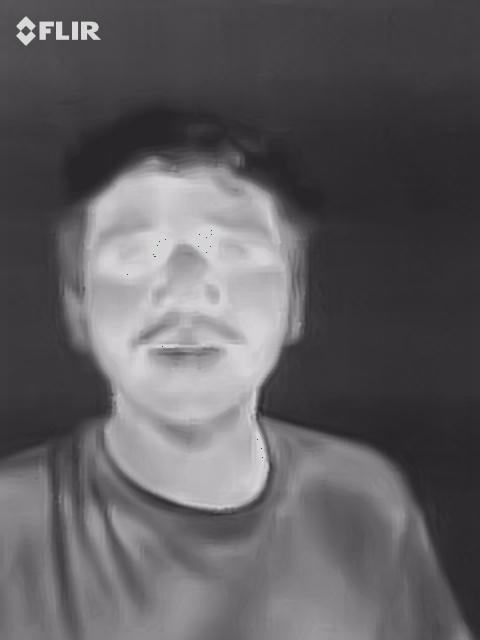}
        \label{fig:13}
    \end{subfigure}
    \begin{subfigure}[b]{0.3\textwidth}
        \includegraphics[width=\textwidth]{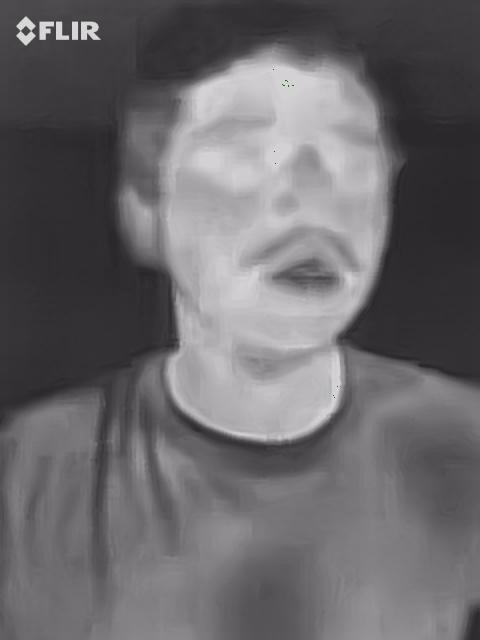}
        \label{fig:10}
    \end{subfigure}
    \caption{Sample images from the dataset}\label{fig:Samples}
\end{figure}

The database presented and used here is multimodal, containing both visible and thermal images that are acquired simultaneously. It contains acquired images of $100$ participants with $10$ image sets of visual and thermal images each at different illumination conditions for the objective of illumination invariant facial recognition. Rather than emphasizing on acquiring data in various modes, we focus on the accuracy of the alignment of sets of visible as well as thermal images. Therefore, our database provides:
\begin{itemize}
\item High-resolution data at $640 \times 480$ pixels, much higher than currently available databases that usually work with $320 \times 240$ pixel data.
\item Facial images at a certain headpose were captured at the same alignment. It assures that they can be superimposed on each other without misalignment in a low light setting. This is possible using a specific setting between the thermal and visible cameras in the hardware.
\item A wide range of head poses instead of the usually fully frontal recordings provided elsewhere. To the best of our knowledge, our database is the only set available with simultaneously aligned facial images in the visual and infrared spectrum at variable illumination.
\end{itemize}
All images for our dataset were recorded using a FLIR One Pro high resolution thermal infrared camera with a $160 \times 120$ pixel-sized microbolometer sensor equipped working in an infrared spectrum range of $8 \mu m-14 \mu m$. Sample images of the created dataset are shown in Figure \ref{fig:Samples}.

\section{Proposed Methodology}

\subsection{Pixel Intensity Based Face Detection Algorithm}
The first step in our proposed pipeline is face detection in thermal images. This algorithm operates based on the pixel intensity profiling of the thermal image, in which the face along with the neck region is extracted. We have considered here two assumptions; one of which is that the facial region is covering the biggest part in the whole image and the other one, pixel intensity of face is largely different from the other parts below neck due to clothing or various reasons. After extraction of the face in the thermal image we can get pre-processed data for the next step which is image fusion to get a fused face. Our proposed method is very simple and does not depend on the detection of any other facial part, a curve of the face or anthropometric relationships but only on the pixel intensity, which is readily available. The steps followed to do so are described in algorithm \ref{alg:euclid}.

\begin{algorithm}
\caption{Face Detection in Infrared Images}\label{alg:euclid}
\begin{algorithmic}[1]
\State A thermal image is obtained in grayscale format as shown in Figure \ref{fig:14c}.
\State Use histogram equalization to improve the contrast of the thermal image. The top 1\% and the bottom 1\% of the pixel values are saturated. (shown in Figure \ref{fig:im1}). 
\State  Use a two dimensional median filter in order to smooth the image. The result (shown in Figure \ref{fig:im2} smooths out the minor discontinuity in the pixel values in a region of the image. 
\State  Apply histogram equalization again to improve the contrast of the image (shown in Figure \ref{fig:im3}. The top 1\% and the bottom 1\% of the pixel values are saturated. 
\State  Now the image we have is multimodal usually having three to four modes. Select the mode having more number of pixels as well as based on the one having bright intensity. For this we threshold the image using the information from the histogram (shown in Figure \ref{fig:im4}. 
\State  Process the thresholded image and to fit the smallest rectangle possible and extract the face region from the original image. (shown in Figure \ref{fig:2018r}). 

\end{algorithmic}
\end{algorithm}

\begin{figure}[H]
 
    \centering
    \begin{subfigure}[b]{0.3\textwidth}
        \includegraphics[width=\textwidth]{14.jpg}
        \caption{}
        \label{fig:14c}
    \end{subfigure}
    \begin{subfigure}[b]{0.3\textwidth}
        \includegraphics[width=\textwidth]{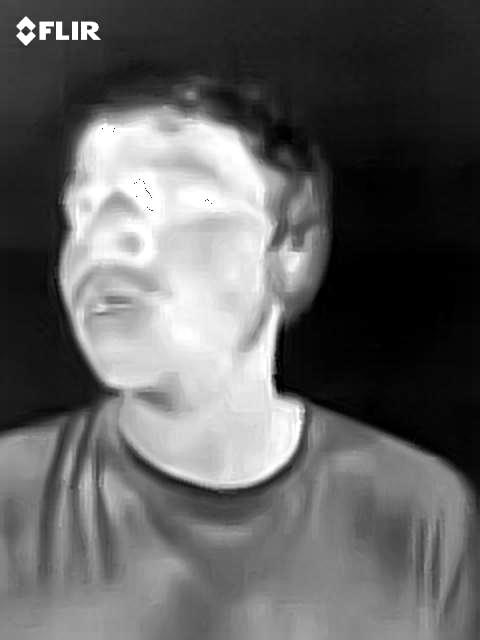}
        \caption{}
        \label{fig:im1}
    \end{subfigure}
    \begin{subfigure}[b]{0.3\textwidth}
        \includegraphics[width=\textwidth]{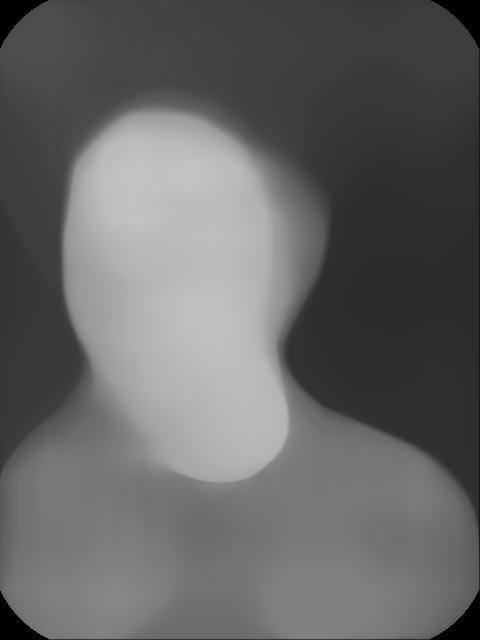}
        \caption{}
        \label{fig:im2}
    \end{subfigure}
    \\
    \begin{subfigure}[b]{0.3\textwidth}
        \includegraphics[width=\textwidth]{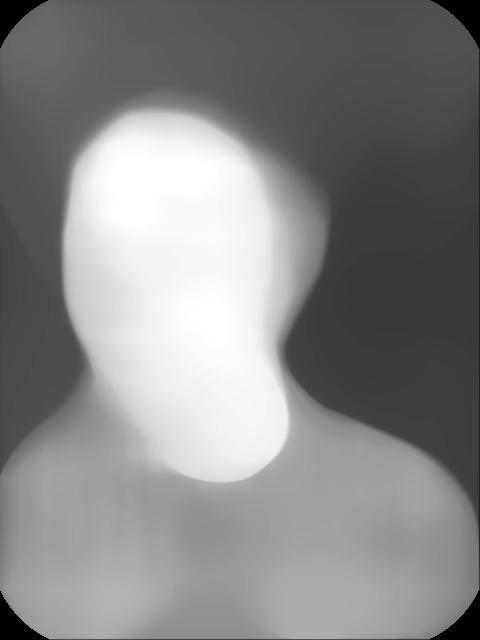}
        \caption{}
        \label{fig:im3}
    \end{subfigure}
    \begin{subfigure}[b]{0.3\textwidth}
        \includegraphics[width=\textwidth]{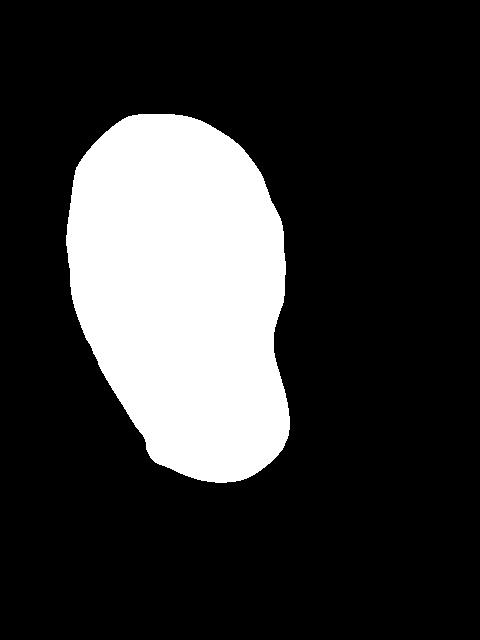}
        \caption{}
        \label{fig:im4}
    \end{subfigure}
    \begin{subfigure}[b]{0.3\textwidth}
        \includegraphics[width=\textwidth]{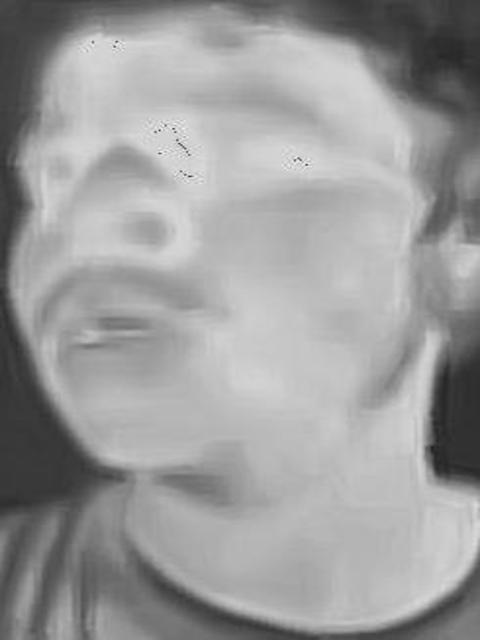}
        \caption{}
        \label{fig:2018r}
    \end{subfigure}
    \caption{Step-wise result of Face Extraction: (a) captured thermal image in grayscale format, (b) Histogram equalized image, (c) result after median filtering, (d) contrast enhanced image, (e) region of our interest and (f) extracted face }\label{fig:FaceResult}
\end{figure}

\subsection{Image Fusion}
After obtaining the corresponding visible and thermal images after the preprocessing step, we work on the lines of Ma et al. \cite{ma2016infrared} present the method of image fusion which uses gradient transfer, and optimizes for the best fusion using total variation minimization. 

\begin{figure}[H]

    \centering
    \begin{subfigure}[b]{0.3\textwidth}
        \includegraphics[width=\textwidth]{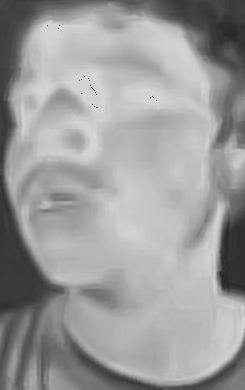}
        \caption{}
        \label{fig:14det}
    \end{subfigure}
    \begin{subfigure}[b]{0.3\textwidth}
        \includegraphics[width=\textwidth]{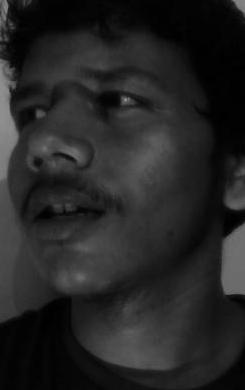}
        \caption{}
        \label{fig:14vis}
    \end{subfigure}
    \begin{subfigure}[b]{0.3\textwidth}
        \includegraphics[width=\textwidth]{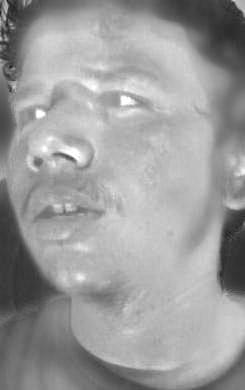}
        \caption{}
        \label{fig:14fus}
    \end{subfigure}
    \\
 
    \caption{Image Fusion Example: (a) Thermal Image, (b)Visual Image representing the same alignment and region as the Thermal Image, (c)Fusion result (Parameters: $\lambda$=7)}\label{fig:Samples-fused}
\end{figure}

Our aim is to create a fused image that preserves both the infrared radiation information and the visual information in the two images given that the images are co-registered and aligned. The example can be seen in Figure \ref{fig:Samples-fused}. Both the visual and thermal images are considered to be grayscale. Let the size of thermal, visible and fused images be $m \times n$ , and their vectors in column forms be denoted by $ir, vi, x, \in R (mn ×1)$, respectively.

Infrared images can reliably distinguish between targets and background by pixel intensity differences. This point provides motivation to force the fused image to have the properties of similar pixel intensities with the given thermal image. This can be achieved by minimizing the empirical error measured by some $ l^r$  norm $( r \geq 1)$ .

\begin{equation}\label{eq1}
\varepsilon_1(\mathnormal{x}) = \frac{1}{r} {{\parallel \mathnormal{x}-\mathnormal {ir} \parallel}^r}_r
\end{equation}

Also, we want the fused image to preserve the characteristics on the visual image. The fused image should have similar pixel intensities with the visual image in order to fuse the detailed appearance information. But, the infrared and visual images represent different phenomena which leads to different pixel intensities in the same pixel location making it unsuitable to generate x by simultaneously minimizing $\frac{1}{r} {{\parallel \nabla \mathnormal{x}-\nabla \mathnormal{ir}\parallel}^r}_r$ and $\frac{1}{s} {{\parallel \nabla \mathnormal{x}-\nabla \mathnormal{vi}\parallel}^s}_s$. 

We consider gradients of the image to characterize the detailed appearance information about the scene. Therefore, we propose to constrain the fused image to have similar pixel gradients rather than similar pixel intensities with the visible image.

\begin{equation}\label{eq2}
\varepsilon_2(\mathnormal{x}) = \frac{1}{s} {{\parallel \nabla \mathnormal{x}-\nabla\mathnormal{vi} \parallel}^s}_s 
\end{equation}

where $\nabla$ is the gradient operator which we will define in details latter. In the case of s = 0 , Eq. \ref{eq2} is defined as $\varepsilon_2(\mathnormal{x}) = {{\parallel \nabla \mathnormal{x}-\nabla \mathnormal{vi}\parallel}}_0$, which equals the number of non-zero entries of $\nabla x -\nabla vi. $
Hence from Eq. \ref{eq1} and Eq. \ref{eq2} , the fusion problem is formulated as minimization of the following objective function:

\begin{equation}\label{eq3}
\begin{aligned}
\varepsilon(\mathnormal{x})& = \varepsilon_1(\mathnormal{x}) + \lambda \varepsilon_2(\mathnormal{x})\\
& = \frac{1}{r} {{\parallel \mathnormal{x}-\mathnormal{ir}\parallel}^r}_r + \lambda \frac{1}{s} {{\parallel \nabla \mathnormal{x}-\nabla \mathnormal{vi}\parallel}^s}_s
\end{aligned}
\end{equation}

where the first term constrains the fused image x to have the similar pixel intensities with the infrared image ir , the second term requires that the fused image x and the visible image vi have the similar gradients at corresponding positions, and $\lambda$ is a positive parameter to control the trade-off between the two terms. The above objective function extent aims to transfer the edges in the visible image onto the corresponding positions in the infrared image. This method increases the fidelity of a standard infrared image by fusing important information from visual image.

\subsection{Optimization    }

The norms $ l^r$ and $ l^s$, are being considered in the objective function Eq. \ref{eq3} . The Gaussian difference between the fused image x and the infrared image ir will lead to $r = 2$ as a natural choice whereas Laplacian or impulsive case will lead to $r = 1$. Specifically, in our problem we expect to keep the thermal radiation information of u , which means that Most entries of $x − ir$ should be zero since the fused image contains the thermal information. Also a small part of the entries could be large due to the purpose of gradient transfer from the visible image vi . This leads to the difference between x and ir to be Laplacian or impulsive rather than Gaussian, i.e. r = 1 . The property of piece-wise smoothness is often exhibited by natural images leading their gradients to be sparse and of large magnitude at the edges. Encouraging sparseness of the gradients leads to minimizing the  $l^0$ norm, i.e., s = 0 . Since the  $ l^0$ norm is NP-hard, leads to an alternative convex relaxation approach to replace  $ l^0$ by  $ l^1$ . The exact recovery of sparse solutions by  $ l^1$ is guaranteed by the restricted isometry property condition. Therefore, we consider minimizing the gradient differences with $ l^1$ norm, i.e. s = 1 , and  $ l^1$ on the gradient is the total variation. Let y = x − vi , the optimization problem Eq. \ref{eq4} can be rewritten as:

\begin{equation}
\centering
\label{eq4}
\begin{aligned}
\mathnormal{y^*}= arg_\mathnormal{y}  \sum_{i-1}^{mn} |{\mathnormal{y}_i}-(\mathnormal{ir}_i -\mathnormal{vi}_i)| + \lambda \mathnormal{J}(\mathnormal{y}) \\
with \quad \mathnormal{J}(\mathnormal{y}) =\sum_{i-1}^{mn} |{\nabla_i \mathnormal{y}}| = \sum_{i-1}^{mn} \sqrt{({\nabla^h_i\mathnormal{y}})^2 + ({\nabla^{\nu}_i\mathnormal{y}})^2},
\end{aligned}
\end{equation}

where $|x|:= \sqrt{{x_1}^2+{x_2}^2}$ for every $x = (x_1, x_2) \in R^2$, $\nabla_i = ({\nabla_i}^h , {\nabla_i}^v) $ denotes the image gradient $\nabla$ at pixel i with ${\nabla}^h$ and ${\nabla}^v$ being linear operators corresponding to the horizontal and vertical first order differences, respectively. More specifically, ${\nabla_i}^h x = x_i − x_{r ( i )}$ and ${\nabla_i}^v x = x_i − x_{b ( i )}$ , where r(i) and b(i) represent the nearest neighbor to the right and below the pixel i . Besides, if pixel i is located in the last row or column, r(i) and b(i) are both set to be i. The objective function Eq. \ref{eq4} is seen to be convex and thus has a global optimal solution. The second term can be considered as a regularization item, which strikes as an appropriate parameter in considering the detailed appearance information in the visual image. The problem Eq. \ref{eq4} is a standard  $ l^1$ -TV minimization problem and can be understood better using the algorithm proposed in \cite{chan2005aspects}.The proposed GTF algorithm is very simple yet efficient. The global optimal solution of the fused image $x^∗$ is then determined by: ${{\mathnormal{x}}^∗}$ =$ y^∗+v$.

\subsection{Face Recognition on fused images }

Convolutional neural networks \cite{lecun2015deep} are designed to process data that come in the form of multiple arrays. These include a colour image composed of three 2D arrays containing pixel intensities in the three colour channels.  The key ideas on which form the foundation on CNN are: local connections, shared weights, pooling and the use of many layers. A typical CNN comprises of: convolutional layers and pooling layers. Units in a convolutional layer are organized in feature maps, within which each unit is connected to local patches in the feature maps of the previous layer through a set of weights called a filters. The result of this local weighted sum is then passed through a non-linearity such as a ReLU. All units in a feature map share the same filter bank. Using this kind of architecture is beneficial as in array data like images, local group of values are correlated, forming distinctive templates that can be easily detected by the filters. Also, using this filters brings about translation invariance. In other words, if a template can appear in one part of the image, it could appear anywhere, hence the idea of units at different locations sharing the same weights and detecting the same pattern in different parts of the array. Mathematically, the filtering operation performed by a feature map is a discrete convolution, hence the name.

CNN's have been proven to be very useful in computer vision since it preserves the spatial relationship between pixels. In many real world applications such as biometrics, the recognition tasks need to be carried away in a timely fashion on a computationally limited platform. Real time and offline inferencing are some of the major challenges in modern biometrics for the purpose of robust finegrain classification and increased reliability in the absence of internet. 
The class of efficient CNN models called MobileNet are small and have low latency on mobile and embedded applications. Two hyperparameters - 'width multiplier' and 'resolution multiplier' dictates the property of resulting architecture.

\subsection{MobileNet Architecture}
The trend to go deeper and more complicated for achieving higher accuracy in CNN's also leads to high memory requirements and slow inference. We use the MobileNet architecture by Howard et al. \cite{Howard2017MobileNetsEC} so as to make the architecture readily deployable on mobile systems. 
A typical convolutional layer works by taking a input of size $H_{F} \times H_{F} \times M$ feature map F and producing a feature map of size $H_{G}\times H_{G} \times N$, where $H_{F}$, $H_{G}$ is the spatial width and height of square input feature map and square output feature map respectively, M is the number of input channels and N is the number of output channels.

The standard convolutional layer is parameterized by convolution kernel K of size $H_{K} \times H_{K} \times M \times N$ where $H_{K}$ is the spatial dimension of the kernel assumed to be square and M is number of input channels and N is the number of
output channels as defined previously.

Standard convolutional neural networks have a typical cost of $H_{K} \cdot H_{K} \cdot M  \cdot N \cdot H_F \cdot H_F $ . In a MobileNet architecture the same convolution operation is divided into two convolutions called the depthwise and the pointwise convolutions. Depthwise convolutions is used to apply single filter per input channel and the pointwise convolution which is a simple $1 \times 1$ convolutions is used to create linear combinations of output of depthwise layer.

This modification reduces the computation cost to $H_K \cdot H_K \cdot M  \cdot H_F \cdot H_F + M \cdot N \cdot H_F \cdot H_F $, which is the sum of computation cost for depthwise and pairwise computations. The reduction in computation is given by:

\begin{equation}
\centering
\label{eq5}
 \dfrac{H_K \cdot H_K \cdot M  \cdot H_F \cdot H_F + M \cdot N \cdot H_F \cdot H_F}{H_{K} \cdot H_{K} \cdot M  \cdot N \cdot H_F \cdot H_F} = \dfrac{1}{N}+\dfrac{1}{H_{K}^2}
\end{equation}   

\begin{figure}[ht]
\centering
\includegraphics[width=0.7\textwidth]{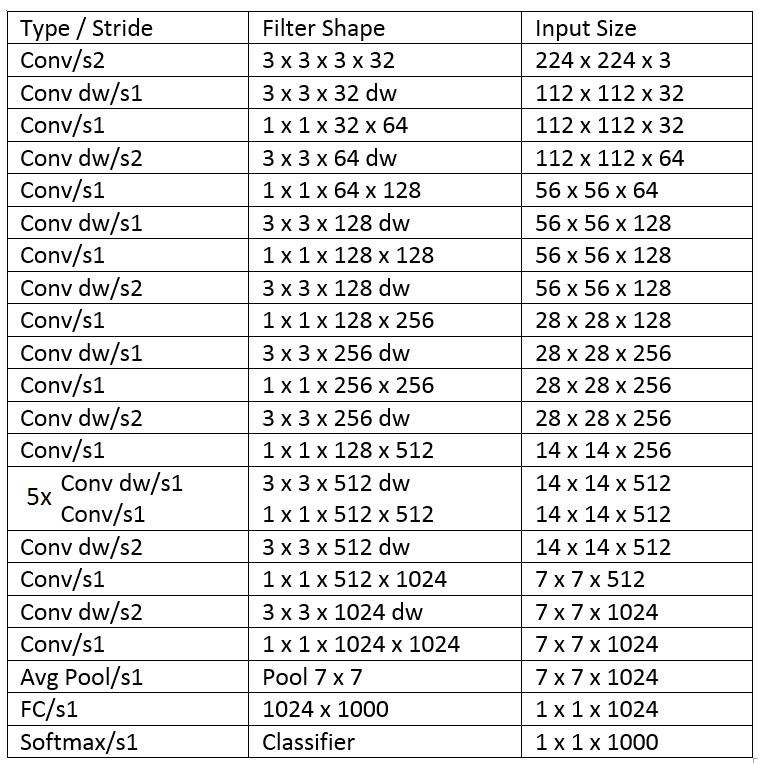}
\caption{Body Architecture of MobileNet}
\label{fig:arrchitecture}
\end{figure}

Eq. \ref{eq5} shows that with a kernel of size $3 \times 3$, we can increase the computation speed by 8 to 9 times than that of standard convolutions with a small reduction in accuracy. The MobileNet architecture can be shown in Figure. \ref{fig:arrchitecture}.

\section{Experiment and Results}

\subsection{ Dataset}

We use our proposed dataset for the face recognition pipeline since there are no well aligned and simultaneous thermal and visual facial image datasets available to our current knowledge. The images are already space alignment registered due to the hardware setting of the visual and thermal lenses in the Flir One camera. There are 10 images at different head poses in each modality per person. The total number of subjects are 100. 

Since there are a limited fused images per person i.e 10 images, dataset augmentation is required in order to increase the generalization capabilities of trained data mobile network. For augmentation, the faces are rotated at 10 degrees from -90 degrees to 90 degrees, thus increasing the dataset size by about 19 times.

\subsection{ Image Fusion parameters}

\begin{figure}[ht]
\centering
\includegraphics[width=0.5\textwidth]{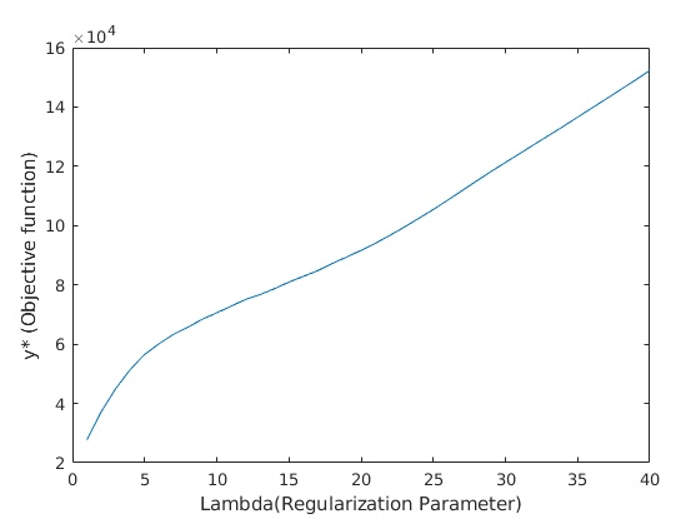}
\caption{Objective function vs Lambda characteristics}
\label{fig:fusionResult}
\end{figure}

As shown in Figure \ref{fig:fusionResult}, the increase in $\lambda$ i.e. regularization parameter leads to increase in the value of optimization function in Eq. \ref{eq4}. Also the increase in  $\lambda$ can be seen as providing more importance to the visual gradient features than the thermal intensity ones in the fused image, leading to the image turning more towards the visual domain. The trade off of both of these properties result in fused images representing varying types of features.

\subsection{ Face Recognition Accuracy}

\begin{figure}[ht]
\centering
\includegraphics[height=6cm,keepaspectratio]{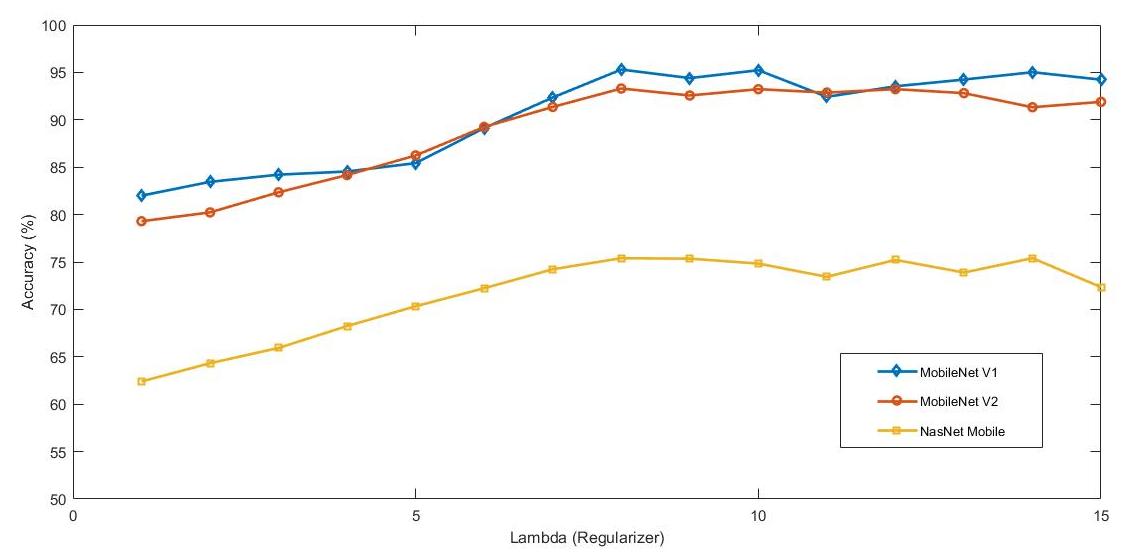}
\caption{Accuracy of recognition of fused images for different models against varying $\lambda$ }
\label{fig:architecture}
\end{figure}

To compare the performance of our proposed framework, we compare our model with two other contemporary deep learning models viz. Nasnet \cite{zoph2017learning} and MobileNet V2 \cite{sandler2018mobilenetv2}. The results of the face recognition for different modes of acquisition  by varying $\lambda$ are shown in Figure. \ref{fig:architecture}. The recognition accuracies for different image modes using different models at $\lambda$=8 are shown in Table \ref{Tab:table1}. Since CNN's perform best for the image data, this leads to better accuracy results for MobileNet since they are the faster and smaller variants of CNN. We can see here that the face recognition for the thermal images provide very less accurate results because of the unavailability of sharp edges i.e. gradient features in the images. During the optimization in the last steps of image fusion the smaller gradient in the visual images are turned into larger values leading to hidden edges in visual images to become more clear in the fused images. The facial heat maps of individuals in the thermal images contain some features unique to the one which when along with the visual features collaborate to give higher accuracy of recognition.

\begin{table}[ht]
\centering
\caption{Face recognition accuracy for image modes by different models at $\lambda$ =8}
\begin{tabular}{|l|c|c|c|}
\hline
 & \multicolumn{1}{l|}{\textbf{Thermal}} & \multicolumn{1}{l|}{\textbf{Visual}} & \multicolumn{1}{l|}{\textbf{Fusion}} \\ \hline
\textbf{NasNet Mobile} & 62.6\% & 66.1\% & 75.4\% \\ \hline
\textbf{MobileNet V2} & 79.3\% & 87.3\% & 93.3\% \\ \hline
\textbf{MobileNet V1} & 82.6\% & 90.7\% & 95.7\% \\ \hline
\end{tabular}
\label{Tab:table1}
\end{table}

\section{Conclusion}
In this paper, we propose a novel methodology for thermal face recognition and experimentally show the superior performance of our approach by evaluating the performance of recognition on the fused images. We create a dataset of both thermal and visual images of faces in simultaneous snaps. This ensures the least error due to the alignment of faces. Visual images have its limitation in scenarios like face spoofing and liveliness detection. Hence we have incorporated the strengths of two modalities i.e. thermal and visual to create a merged representation of the face. We further deploy MobileNet to extract robust features from the merged face to show higher accuracy of face recognition. In the paper, we have shown that the performance of accuracy for the merged face is better than face images in individual modalities. Also, we have captured the face data with certain constraints like the pose, expression, and distance from the camera. In the future, we propose to extend the dataset for the unconstrained environment. We further open the scope for research fraternity to address the face recognition in light of merged modality representation.

\bibliographystyle{IEEEtran}
\bibliography{sample}

\begin{thebibliography}{10}
\providecommand{\url}[1]{#1}
\csname url@samestyle\endcsname
\providecommand{\newblock}{\relax}
\providecommand{\bibinfo}[2]{#2}
\providecommand{\BIBentrySTDinterwordspacing}{\spaceskip=0pt\relax}
\providecommand{\BIBentryALTinterwordstretchfactor}{4}
\providecommand{\BIBentryALTinterwordspacing}{\spaceskip=\fontdimen2\font plus
\BIBentryALTinterwordstretchfactor\fontdimen3\font minus
  \fontdimen4\font\relax}
\providecommand{\BIBforeignlanguage}[2]{{%
\expandafter\ifx\csname l@#1\endcsname\relax
\typeout{** WARNING: IEEEtran.bst: No hyphenation pattern has been}%
\typeout{** loaded for the language `#1'. Using the pattern for}%
\typeout{** the default language instead.}%
\else
\language=\csname l@#1\endcsname
\fi
#2}}
\providecommand{\BIBdecl}{\relax}
\BIBdecl

\bibitem{ekenel2007face}
H.~K. Ekenel, J.~Stallkamp, H.~Gao, M.~Fischer, and R.~Stiefelhagen, ``Face
  recognition for smart interactions,'' in \emph{Multimedia and Expo, 2007 IEEE
  International Conference on}.\hskip 1em plus 0.5em minus 0.4em\relax IEEE,
  2007, pp. 1007--1010.

\bibitem{pentland2000face}
A.~Pentland and T.~Choudhury, ``Face recognition for smart environments,''
  \emph{Computer}, vol.~33, no.~2, pp. 50--55, 2000.

\bibitem{galton1889personal}
F.~Galton, ``Personal identification and description,'' \emph{Journal of
  Anthropological Institute of Great Britain and Ireland}, pp. 177--191, 1889.

\bibitem{belhumeur1997eigenfaces}
P.~N. Belhumeur, J.~P. Hespanha, and D.~J. Kriegman, ``Eigenfaces vs.
  fisherfaces: Recognition using class specific linear projection,'' Yale
  University New Haven United States, Tech. Rep., 1997.

\bibitem{he2005face}
X.~He, S.~Yan, Y.~Hu, P.~Niyogi, and H.-J. Zhang, ``Face recognition using
  laplacianfaces,'' \emph{IEEE transactions on pattern analysis and machine
  intelligence}, vol.~27, no.~3, pp. 328--340, 2005.

\bibitem{gao2002face}
Y.~Gao and M.~K. Leung, ``Face recognition using line edge map,'' \emph{IEEE
  Transactions on Pattern Analysis \& Machine Intelligence}, no.~6, pp.
  764--779, 2002.

\bibitem{kirby1990application}
M.~Kirby and L.~Sirovich, ``Application of the karhunen-loeve procedure for the
  characterization of human faces,'' \emph{IEEE Transactions on Pattern
  analysis and Machine intelligence}, vol.~12, no.~1, pp. 103--108, 1990.

\bibitem{bartlett2002face}
M.~S. Bartlett, J.~R. Movellan, and T.~J. Sejnowski, ``Face recognition by
  independent component analysis,'' \emph{IEEE transactions on neural
  networks/a publication of the IEEE Neural Networks Council}, vol.~13, no.~6,
  p. 1450, 2002.

\bibitem{adini1997face}
Y.~Adini, Y.~Moses, and S.~Ullman, ``Face recognition: The problem of
  compensating for changes in illumination direction,'' \emph{IEEE Transactions
  on pattern analysis and machine intelligence}, vol.~19, no.~7, pp. 721--732,
  1997.

\bibitem{wen2015face}
D.~Wen, H.~Han, and A.~K. Jain, ``Face spoof detection with image distortion
  analysis,'' \emph{IEEE Transactions on Information Forensics and Security},
  vol.~10, no.~4, pp. 746--761, 2015.

\bibitem{cutler1996face}
R.~G. Cutler, \emph{Face recognition using infrared images and
  eigenfaces}.\hskip 1em plus 0.5em minus 0.4em\relax University of Maryland,
  1996.

\bibitem{bebis2006face}
G.~Bebis, A.~Gyaourova, S.~Singh, and I.~Pavlidis, ``Face recognition by fusing
  thermal infrared and visible imagery,'' \emph{Image and Vision Computing},
  vol.~24, no.~7, pp. 727--742, 2006.

\bibitem{socolinsky2003face}
D.~A. Socolinsky, A.~Selinger, and J.~D. Neuheisel, ``Face recognition with
  visible and thermal infrared imagery,'' \emph{Computer vision and image
  understanding}, vol.~91, no. 1-2, pp. 72--114, 2003.

\bibitem{forczmanski2016human}
P.~Forczma{\'n}ski, ``Human face detection in thermal images using an ensemble
  of cascading classifiers,'' in \emph{International Multi-Conference on
  Advanced Computer Systems}.\hskip 1em plus 0.5em minus 0.4em\relax Springer,
  2016, pp. 205--215.

\bibitem{wong2012face}
W.~K. Wong, J.~H. Hui, J.~B.~M. Desa, N.~I. N.~B. Ishak, A.~B. Sulaiman, and
  Y.~B.~M. Nor, ``Face detection in thermal imaging using head curve
  geometry,'' in \emph{Image and Signal Processing (CISP), 2012 5th
  International Congress on}.\hskip 1em plus 0.5em minus 0.4em\relax IEEE,
  2012, pp. 881--884.

\bibitem{selinger2006appearance}
A.~Selinger and D.~A. Socolinsky, ``Appearance-based facial recognition using
  visible and thermal imagery: a comparative study,'' EQUINOX CORP NEW YORK NY,
  Tech. Rep., 2006.

\bibitem{nguyen2013thermal}
H.~Nguyen, K.~Kotani, F.~Chen, and B.~Le, ``A thermal facial emotion database
  and its analysis,'' in \emph{Pacific-Rim Symposium on Image and Video
  Technology}.\hskip 1em plus 0.5em minus 0.4em\relax Springer, 2013, pp.
  397--408.

\bibitem{wang2010natural}
S.~Wang, Z.~Liu, S.~Lv, Y.~Lv, G.~Wu, P.~Peng, F.~Chen, and X.~Wang, ``A
  natural visible and infrared facial expression database for expression
  recognition and emotion inference,'' \emph{IEEE Transactions on Multimedia},
  vol.~12, no.~7, pp. 682--691, 2010.

\bibitem{ma2016infrared}
J.~Ma, C.~Chen, C.~Li, and J.~Huang, ``Infrared and visible image fusion via
  gradient transfer and total variation minimization,'' \emph{Information
  Fusion}, vol.~31, pp. 100--109, 2016.

\bibitem{chan2005aspects}
T.~F. Chan and S.~Esedoglu, ``Aspects of total variation regularized l 1
  function approximation,'' \emph{SIAM Journal on Applied Mathematics},
  vol.~65, no.~5, pp. 1817--1837, 2005.

\bibitem{lecun2015deep}
Y.~LeCun, Y.~Bengio, and G.~Hinton, ``Deep learning,'' \emph{nature}, vol. 521,
  no. 7553, p. 436, 2015.

\bibitem{Howard2017MobileNetsEC}
A.~G. Howard, M.~Zhu, B.~Chen, D.~Kalenichenko, W.~Wang, T.~Weyand,
  M.~Andreetto, and H.~Adam, ``Mobilenets: Efficient convolutional neural
  networks for mobile vision applications,'' \emph{CoRR}, vol. abs/1704.04861,
  2017.

\bibitem{zoph2017learning}
B.~Zoph, V.~Vasudevan, J.~Shlens, and Q.~V. Le, ``Learning transferable
  architectures for scalable image recognition,'' \emph{arXiv preprint
  arXiv:1707.07012}, vol.~2, no.~6, 2017.

\bibitem{sandler2018mobilenetv2}
M.~Sandler, A.~Howard, M.~Zhu, A.~Zhmoginov, and L.-C. Chen, ``Mobilenetv2:
  Inverted residuals and linear bottlenecks,'' in \emph{Proceedings of the IEEE
  Conference on Computer Vision and Pattern Recognition}, 2018, pp. 4510--4520.

\end{thebibliography}

\end{document}